\documentclass[unnumsec,webpdf,contemporary,large]{oup-authoring-template}
\usepackage{epstopdf}
\usepackage{graphicx}
\usepackage{xcolor}
\graphicspath{{figures/}}
\DeclareRobustCommand{\rev}[1]{\textcolor{black}{#1}}

\theoremstyle{thmstyleone}%
\theoremstyle{thmstyletwo}%
\theoremstyle{thmstylethree}%

\begin{document}

\journaltitle{Preprint}
\DOI{Preprint version}
\copyrightyear{2026}
\pubyear{2026}
\access{Preprint posted: September 2026}
\appnotes{Preprint}

\firstpage{1}


\title{ProMeta: Few-shot PROTAC-targeted degradation prediction across E3 ligases}

\author{
Yuansheng Liu$^{1,2}$,
Yufei Ye$^{1,2}$,
Tao Tang$^{3}$,
Jiawei Luo$^{1,2}$,
Wen Tao$^{4,*}$,
Xiao Luo$^{5,*}$
}

\address{
$^{1}$College of Computer Science and Electronic Engineering, Hunan University, Changsha, Hunan 410086, China\\
$^{2}$Yuelushan Laboratory, Changsha, Hunan 410128, China\\
$^{3}$School of Modern Posts, Nanjing University of Posts and Telecommunications, Nanjing, Jiangsu 210023, China\\
$^{4}$College of Computing and Data Science, Nanyang Technological University, Singapore 639798, Singapore\\
$^{5}$Hunan Research Center of the Basic Discipline for Cell Signaling, College of Biology, Hunan University, Changsha, Hunan 410086, China
}

\corresp{*Corresponding author: taowen228@gmail.com; xluo@hnu.edu.cn}

\abstract{
\textbf{Motivation:} Proteolysis-targeting chimeras (PROTACs) have emerged as a transformative therapeutic strategy that selectively degrades historically ``undruggable'' targets via the ubiquitin–proteasome system. Despite growing efforts to develop computational predictors of PROTAC degradation activity, existing supervised approaches remain severely challenged by data scarcity and imbalance across E3 ligases, limiting their ability to generalize beyond well-studied ligase contexts. In practice, labeled data are heavily concentrated on a few ligases (e.g., CRBN and VHL), while the majority of E3 ligases remain underexplored yet are critical for expanding the design space of targeted degraders. Developing methods that enable robust cross-ligase generalization with minimal labeled data is therefore essential for improving the practical utility of computational PROTAC discovery. \newline
\textbf{Results:} We reformulate PROTAC degradation activity prediction across E3 ligases as a few-shot meta-learning problem and present ProMeta, a prototype-based graph neural network trained through episodic meta-learning on source-E3 tasks and evaluated on held-out target-E3 tasks through support-conditioned inference. ProMeta performs inference without updating the encoder by dynamically estimating class prototypes from minimal target-ligase support samples. \rev{On the CRBN-to-VHL benchmark, ProMeta achieves AUROC values of $0.796$ under $K=2, Q=3$ and $0.883$ under $K=2, Q=5$, improving by 19.9\% and 6.8\%, respectively, over the corresponding supervised GNN baseline. }\rev{Reverse VHL-to-CRBN transfer under the same protocol yielded
AUROC values of $0.702$ ($K=2, Q=3$) and $0.821$ ($K=2, Q=5$),
confirming bidirectional applicability while revealing direction and data-regime dependence.} \rev{Under one-shot evaluation on rare E3 ligases, ProMeta attains an AUROC of $0.700$.} Case-study analyses further demonstrate predictive consistency on VZ185-derived candidates and two additional retrospective chemical series. Together, these results support ProMeta as a practical framework for cross-ligase few-shot prediction under the evaluated support/query protocols. \newline
\textbf{Availability and implementation:} The source code is available at \url{https://github.com/yeyufeiyyf/prometa}, \rev{and the datasets, experimental splits, model checkpoints, and released prediction results are archived on Zenodo at \url{https://doi.org/10.5281/zenodo.21371599}.}
}
\maketitle

\section{Introduction}

Proteolysis-targeting chimeras (PROTACs) have emerged as a promising therapeutic strategy for targeted protein degradation \cite{Sakamoto2001, Bondeson2015}. Unlike conventional small-molecule inhibitors that rely on occupancy-driven mechanisms, PROTACs recruit an E3 ubiquitin ligase to a protein of interest (POI), inducing ternary complex formation and subsequent ubiquitination and proteasomal degradation \cite{Winter2015}. Operating through an event-driven pharmacological mechanism, PROTACs act catalytically, enabling sustained target depletion at sub-stoichiometric doses and potentially reducing side effects \cite{Toure2016, Neklesa2017}. Importantly, this paradigm allows engagement of proteins traditionally considered ``undruggable'' \cite{Burslem2017, Bekesetal2022}. Collectively, these properties position PROTACs as a transformative modality with the potential to fundamentally reshape therapeutic strategies for challenging protein targets.

Despite their considerable promise, the rational design of effective PROTAC molecules remains challenging. A typical PROTAC comprises three components: a warhead targeting the protein of interest, a ligand recruiting an E3 ubiquitin ligase, and a linker connecting the two. \rev{Degradation efficacy depends on ligand binding affinity and linker length and flexibility \cite{Bekesetal2022}; productive degradation further depends on the cooperativity and structural compatibility of ternary-complex recognition \cite{Gadd2017}.} Experimental screening of large PROTAC libraries is both time-consuming and costly \cite{Hu2019, Kao2023}. With the increasing availability of curated datasets such as PROTAC-DB \cite{Ge2025}, researchers have explored machine learning approaches for activity prediction. Structure-aware models, including graph neural networks (GNNs) \cite{Gilmer2017}, DeepPROTAC \cite{Li2022}, and PROTAC-STAN \cite{chen2025}, as well as multimodal frameworks such as DegradeMaster \cite{liu2025}, have demonstrated encouraging performance by capturing molecular and protein-level interactions. \rev{A recent geometry-aware predictor, SE(3)-PROTACs, combines an SE(3)-equivariant molecular encoder with POI and E3 protein-sequence representations \cite{Kothakapu2026}.} \rev{Nevertheless, whether these prediction frameworks transfer to heterogeneous, label-scarce cross-ligase settings remains insufficiently established.}

Despite promising progress, existing approaches face critical limitations. Most models rely heavily on large annotated datasets and exhibit limited generalization across E3 ligases, target proteins, and chemical scaffolds, while real-world PROTAC data are typically scarce and highly imbalanced. \rev{In our raw PROTAC-DB export, over 80\% of entries lack degradation activity annotations (DC$_{50}$ and D$_{\max}$; data source: PROTAC-DB 3.0 \cite{Ge2025}), and the labeled data used here are predominantly concentrated in CRBN and VHL (Supplementary Fig.~S1). This concentration motivates, but does not itself demonstrate, the challenge posed by rare-ligase settings.} Therefore, developing methods capable of robust cross-ligase generalization under minimal labeled data is critical for enabling data-efficient exploration of underrepresented ligase contexts.

To address this challenge, we reformulate PROTAC degradation activity prediction across E3 ligases as a few-shot meta-learning problem. \rev{This formulation follows the support/query episodic paradigm used by Matching Networks \cite{Vinyals2016} and Prototypical Networks \cite{Snell2017}, while differing from gradient-based fast-adaptation methods such as MAML \cite{Finn2017}; broader meta-learning taxonomies are reviewed by Hospedales et al.\ \cite{Hospedales2022}.} \rev{Each E3-ligase-defined prediction context is treated as a task. E3 identity defines the meta-training/meta-test task split: source-E3 episodes are used for episodic meta-training, whereas held-out target-E3 tasks are instantiated at meta-test from labeled support compounds and evaluated on disjoint query compounds. The class labels remain active/inactive within each task, and target-E3 query labels are never used for prototype construction or model fitting.} Most ligases beyond CRBN and VHL have only limited labeled samples available, and the objective is to generalize to unseen ligases given only a small support set. This setting departs from conventional supervised learning, which assumes a shared data distribution and sufficient annotations, and instead requires learning transferable representations that remain useful under a ligase-defined task shift.

Motivated by this perspective, we propose ProMeta, a prototype-based graph neural network trained with episodic meta-learning. ProMeta encodes each molecule as a graph and augments the molecular representation with protein sequence embeddings of both the POI and the E3 ligase, providing biologically informed context. At inference, this learned representation is complemented by a fixed ECFP4 molecular fingerprint before prototype construction. By learning across diverse source-E3 episodes, the model is optimized for support-conditioned classification on held-out E3 tasks. Class prototypes are then estimated on the fly from a minimal target-ligase support set, enabling prediction for held-out E3 ligases without encoder updates. \rev{We evaluate ProMeta on cross-ligase transfer benchmarks, including bidirectional transfer between CRBN and VHL and an exploratory extreme low-data evaluation across rare E3 ligases, supporting ProMeta as a practical framework for data-scarce PROTAC degradation prediction.}

\section{Materials and Methods}\label{sec:Methods}

\subsection{Dataset collection and preprocessing}

\textbf{Data source and activity metrics.}
To evaluate the proposed framework and all baseline methods, we utilized data from PROTAC-DB~\cite{Ge2025}, a publicly available repository of experimentally characterized PROTAC molecules. The raw local PROTAC-DB export used in this study contains 9,384 rows; this approximately 9.38K-row raw export should be distinguished from the smaller filtered labeled tables used for model training and evaluation. Records provide molecular structures, POI and E3 annotations, and, where available, degradation evidence including DC$_{50}$ and D$_{\max}$. These metrics serve as standard quantitative measures of degradation efficiency~\cite{Pettersson2019, Liu2022}: a lower DC$_{50}$ indicates greater potency, and a higher D$_{\max}$ indicates more complete target elimination.

\textbf{Binary labeling and data filtering.}
Binary activity labels were assigned following the protocol of Li et al.~\cite{Li2022}, incorporating both quantitative DC$_{50}$/D$_{\max}$ values and qualitative activity calls derived from experimental descriptions. For rows with complete quantitative evidence, a compound was designated as \textit{high degradation activity} when DC$_{50}<100$~nM and D$_{\max}\geq80\%$; qualitative-only rows were retained only when an explicit activity call was available. Other retained labeled rows were assigned to the \textit{low degradation activity} class. Raw rows without usable quantitative measurements or qualitative activity calls were excluded rather than treated as negative examples. A high-quality labeled subset was curated through three sequential filtering steps: (i) entries with missing SMILES, UniProt IDs, or activity labels were discarded; (ii) all structures were validated with RDKit~\cite{RDKit}; and (iii) duplicates were removed via canonical SMILES comparison. For the strict CRBN/VHL transfer experiments, duplicate compound--E3--target contexts with conflicting binary labels were removed before support/query split generation, yielding 1,386 final clean CRBN/VHL contexts (842 CRBN and 544 VHL; 690 positive and 696 negative labels).

\textbf{Class imbalance adjustment.}
The raw export exhibits pronounced imbalance across E3 ligases: CRBN (6,041 rows) and VHL (2,858 rows) account for most records before strict activity-evidence filtering (Supplementary Fig.~S1). To mitigate class imbalance in the labeled modeling table, random majority-class down-sampling~\cite{He2009} was applied independently to CRBN and VHL. The balanced labeled dataset shown in Supplementary Fig.~S1 and Table~\ref{tab:e3_distribution} comprises 860 CRBN samples (430 high-activity / 430 low-activity) and 560 VHL samples (280 high-activity / 280 low-activity). This distribution summary precedes the additional context-level duplicate/conflict cleanup that produced the stricter 1,386-context CRBN/VHL analysis set used for the main transfer experiments (Supplementary Table~S1).

\textbf{Protein sequence retrieval.} For each compound in the final dataset, the amino acid sequences of both the POI and the recruited E3 ligase were retrieved from UniProt. POI sequences were obtained using the provided UniProt identifiers, with 89 entries failing to map to a valid sequence and subsequently removed. E3 ligase sequences were retrieved via a manually curated name-to-UniProt mapping covering all ligases present in the dataset. Sequences were truncated to a maximum length of 2,000 residues. To clarify how the final analysis set was obtained without overloading the main text, Supplementary Table~S1 summarizes the preprocessing flow from the raw PROTAC-DB export to the strict CRBN/VHL context-level dataset used for the main transfer experiments. The 1,420-record CRBN/VHL subset is reported only as the candidate pool before duplicate/conflict cleanup; the main strict CRBN/VHL transfer experiments use the final 1,386 unique contexts.

\begin{table}[tbp]
\centering
\caption{Sample distribution across major E3 ligases in the balanced labeled dataset. Ligases not listed individually are pooled as ``Others''.}
\label{tab:e3_distribution}
\begin{tabular}{lcccc}
\toprule
E3 ligase & Low-activity & High-activity & Total  \\
\midrule
CRBN   & 430 & 430 & 860  \\
VHL    & 280 & 280 & 560  \\
cIAP1  &   8 &   8 &  16  \\
IAP    &   7 &   7 &  14  \\
XIAP   &   3 &   3 &   6  \\
FEM1B  &   2 &   2 &   4  \\
MDM2   &   2 &   2 &   4  \\
Others & 129 &   8 & 137  \\
\botrule
\end{tabular}
\end{table}

\subsection{Episodic meta-learning dataset construction}

\textbf{Ligase-disjoint data partitioning.}
To ensure unbiased assessment of cross-ligase generalization, we adopted a ligase-disjoint episodic data partition. \rev{For CRBN$\to$VHL and rare-E3 transfer, CRBN-associated molecules formed the source pool; for the reverse VHL$\to$CRBN benchmark, VHL-associated molecules formed the source pool. Each source pool was split into training (80\%) and validation (20\%) partitions, and the target ligase was excluded from both. The rare-E3 targets were cIAP1, IAP, MDM2, XIAP, and FEM1B.} Ligases without both positive and negative samples were excluded. No compound overlap existed between source and target partitions, and all experiments were conducted under three random seeds (42, 2025, 3407).

\textbf{Benchmark design and justification.}
CRBN-recruiting and VHL-recruiting PROTACs employ chemically distinct ligands that occupy non-overlapping regions of chemical space. Their POI targets are also largely non-overlapping, ensuring the CRBN-to-VHL benchmark reflects genuine cross-distribution transfer. \rev{We selected CRBN as the primary source E3 ligase because it is the most data-rich ligase in PROTAC-DB, reflecting a realistic transfer from a data-rich source to held-out target-ligase contexts. We additionally evaluated the reverse VHL-to-CRBN direction under the same ligase-disjoint episodic protocol as a complementary test of direction dependence. In each transfer direction, the target ligase was excluded from source training and used only through labeled support compounds for prototype construction, while query compounds were reserved exclusively for evaluation.}

\subsection{Framework overview}
\begin{figure*}[t]
\centering
\includegraphics[width=\textwidth]{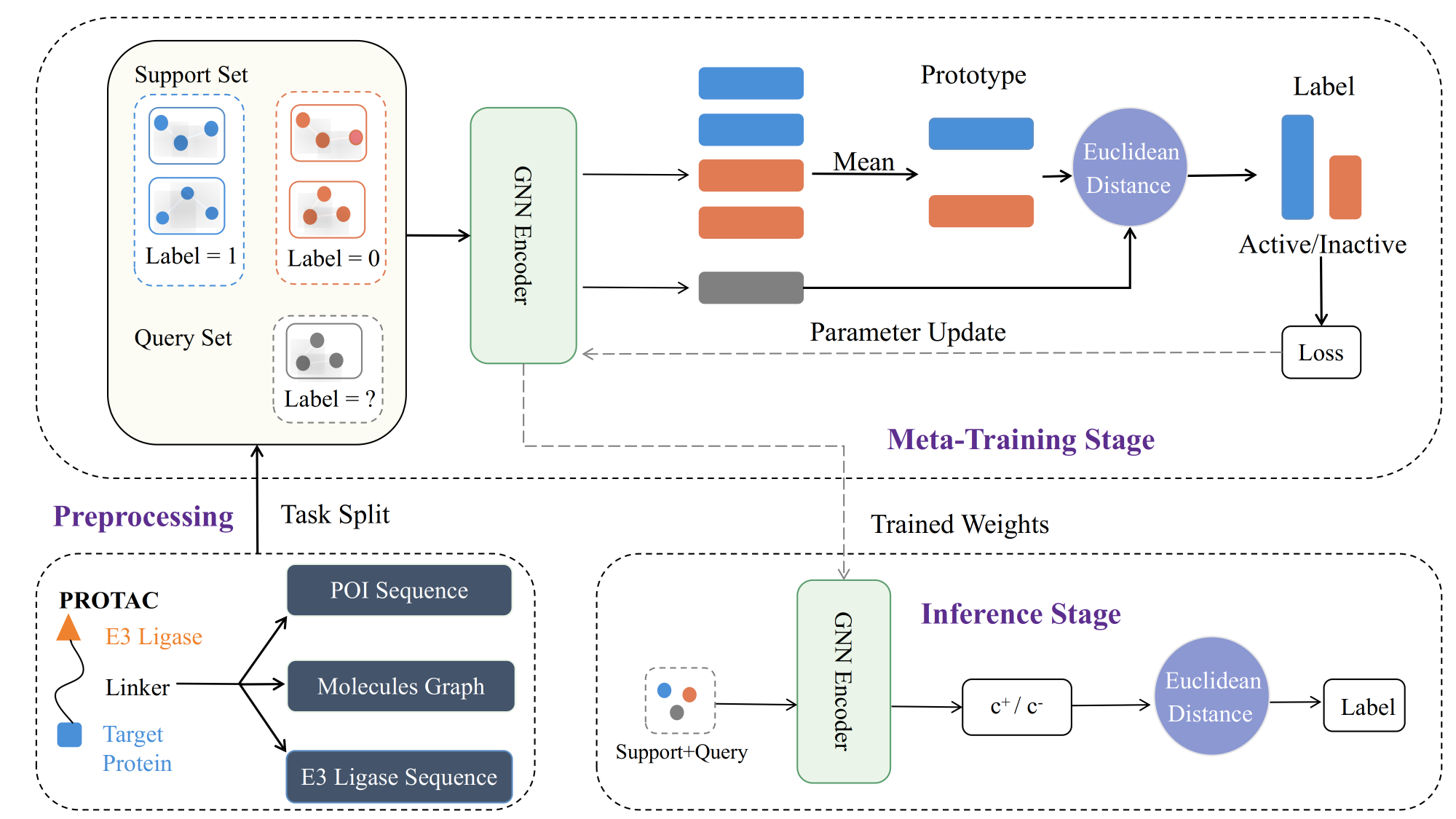}
\caption{Overview of ProMeta, a prototype-based few-shot meta-learning framework for cross-ligase PROTAC degradation activity prediction. PROTAC molecules are constructed as molecular graphs from SMILES strings via RDKit. During episodic meta-training on a source E3, a shared GNN encoder maps molecules into an embedding space, from which class prototypes are computed as the mean embeddings of support-set molecules per class. Query molecules are classified by distance to the prototypes, and encoder parameters are updated through the episodic query loss. At cross-ligase meta-test, the encoder is frozen and transferred to a held-out target E3, enabling support-conditioned prediction without encoder updates. Before prototypes and query distances are computed, the L2-normalized learned representation is concatenated with a weighted L2-normalized 512-bit ECFP4 fingerprint.}
\label{fig:model}

\end{figure*}

An overview of the architecture of ProMeta and its key components is illustrated in Fig.~\ref{fig:model}. As depicted in Fig.~\ref{fig:model}, ProMeta integrates three principal components: a molecular representation module combining learned graph features \cite{Gilmer2017} with protein features and a fixed molecular fingerprint at inference, a prototype-based classifier, and an episodic meta-learning training scheme \cite{Snell2017}. \rev{Consistent with prototypical few-shot learning, the encoder is meta-trained on source-E3 episodes, whereas cross-ligase meta-testing on a held-out target E3 is performed by constructing prototypes from that ligase's support set.} Class prototypes are then estimated from a small labeled support set, and a query molecule is assigned to the class whose prototype is nearest in the learned representation space. This design confers two key advantages over conventional supervised classifiers: (i)~the episodic meta-training objective directly matches the few-shot inference setting, promoting the acquisition of broadly transferable molecular representations; and (ii)~the classification mechanism is non-parametric and requires no parameter update when the target E3 ligase changes, enabling immediate deployment to newly characterized ligase contexts.

\subsubsection{Molecular graph representation}
\textbf{Graph construction.} Each PROTAC molecule is represented as an attributed graph $G = (V, E)$, where nodes $v \in V$ correspond to heavy atoms and edges $e \in E$ to covalent bonds. Graphs were constructed from SMILES strings using RDKit \cite{RDKit}. Node features are encoded as a 9-dimensional one-hot vector representing the atom type, covering the nine most prevalent elements in PROTAC structures: C, N, O, F, P, S, Cl, Br, and I. Edges are included for all covalent bonds and stored as undirected pairs (two directed edges per bond); no edge features are used. All graphs were validated prior to GNN input, and molecules that failed RDKit parsing were excluded during preprocessing.

\noindent\textbf{GNN encoder.} We adopt a Graph Convolutional Network (GCN)~\cite{Kipf2016} as the molecular encoder. Let $h_v^{(l)} \in \mathbb{R}^d$ denote the representation of node $v$ at layer $l$, initialized with the atom feature vector ($l=0$). Each layer updates node representations through neighborhood aggregation:
\begin{equation}
h_v^{(l+1)} = \phi\!\left(h_v^{(l)},\;
  \mathrm{AGG}\!\left(\bigl\{h_u^{(l)} : u \in \mathcal{N}(v)\bigr\}\right)\right),
\label{eq:mpnn}
\end{equation}
where $\mathcal{N}(v)$ denotes the set of atoms bonded to $v$, $\mathrm{AGG}(\cdot)$ is a permutation-invariant aggregation function (mean pooling in this work), and $\phi(\cdot)$ is a linear transformation followed by ReLU activation. Stacking $L$ such layers integrates structural information from the $L$-hop neighborhood of each atom.

A graph-level representation is obtained by global mean pooling over all final node embeddings:
\begin{equation}
\mathbf{z} = \frac{1}{|V|}\sum_{v \in V} h_v^{(L)} \in \mathbb{R}^d,
\label{eq:readout}
\end{equation}
where $\mathbf{z}$ serves as the molecular embedding for all downstream operations. Mean pooling was used as a size-normalizing design choice to reduce sensitivity to molecular size, which varies considerably across PROTAC compounds.

\noindent\textbf{Protein sequence encoding and feature fusion.}
To incorporate biological context, the molecular embedding $\mathbf{z}$ is augmented with protein sequence embeddings of the POI and the E3 ligase. For each compound, the amino acid sequence is encoded by a lightweight sequence encoder consisting of a learnable embedding layer over the 20 standard amino acids, followed by mean pooling across residue positions and a linear projection into $\mathbb{R}^d$. Let $\mathbf{z}^{\text{poi}}$ and $\mathbf{z}^{\text{e3}}$ denote the resulting sequence embeddings for the POI and E3 ligase, respectively. The fused molecular representation is defined as:
\begin{equation}
\tilde{\mathbf{z}} = \mathbf{z} + \alpha_{\text{poi}}\,\mathbf{z}^{\text{poi}} + \alpha_{\text{e3}}\,\mathbf{z}^{\text{e3}},
\label{eq:fusion}
\end{equation}
where $\alpha_{\text{poi}}$ and $\alpha_{\text{e3}}$ are learnable scalar weights initialized to 0.1, allowing the model to adaptively balance molecular structure and biological sequence signals. The fused embedding $\tilde{\mathbf{z}}$ forms the learned molecular/protein block of the representation used for prototype construction.

\noindent\textbf{Molecular fingerprint fusion.}
To complement the learned representation with a fixed descriptor of local chemical substructures, we used RDKit~\cite{RDKit} to generate a 512-bit extended-connectivity fingerprint with diameter four (ECFP4; Morgan radius~$=2$)~\cite{Rogers2010} from each molecule's SMILES string. Let $\mathbf{f}_i$ denote the fingerprint of molecule $i$. At inference, the learned molecular/protein embedding and fingerprint are independently L2-normalized and concatenated as
\begin{equation}
\hat{\mathbf{z}}_i =
\left[
\frac{\tilde{\mathbf{z}}_i}{\lVert\tilde{\mathbf{z}}_i\rVert_2}
\,\middle\|\,
\lambda_{\mathrm{fp}}\frac{\mathbf{f}_i}{\lVert\mathbf{f}_i\rVert_2}
\right],
\label{eq:fingerprint_fusion}
\end{equation}
where $\|$ denotes concatenation and $\lambda_{\mathrm{fp}}=0.5$ controls the contribution of the fingerprint block. The fingerprint is fixed rather than learned and is introduced only when forming support and query features for prototype-based inference on the target ligase.

\subsubsection{Prototype-based classification}
For an episodic task with support set $\mathcal{S} = \{(\mathbf{x}_i, y_i)\}$, the prototype of class $c \in \{0,1\}$ is defined as the centroid of its support-set embeddings:
\begin{equation}
\mathbf{p}_c = \frac{1}{|S_c|}\sum_{(\mathbf{x}_i,\,y_i)\in S_c} f(\mathbf{x}_i),
\label{eq:prototype}
\end{equation}
where $S_c = \{(\mathbf{x}_i,y_i)\in\mathcal{S}: y_i=c\}$. The representation $f(\mathbf{x}_i)$ is $\tilde{\mathbf{z}}_i$ during episodic meta-training and the fingerprint-augmented representation $\hat{\mathbf{z}}_i$ (Eq.~\ref{eq:fingerprint_fusion}) during target-ligase meta-test inference. For a query molecule $\mathbf{x}_q$, the Euclidean distance to each prototype is computed as:
\begin{equation}
d_c = \bigl\|f(\mathbf{x}_q) - \mathbf{p}_c\bigr\|_2,
\label{eq:distance}
\end{equation}
and class assignment probabilities are obtained via a softmax over negative distances, where $c'$ denotes all classes in $\{0,1\}$:
\begin{equation}
P(y=c\mid\mathbf{x}_q) = 
  \frac{\exp(-d_c)}{\sum_{c'}\exp(-d_{c'})}.
\label{eq:softmax}
\end{equation}
Euclidean distance was selected over cosine similarity because it directly reflects absolute displacement in the embedding space, which is more meaningful when prototype positions carry geometric significance. During inference, encoder weights are frozen, and prototypes are recomputed solely from the available support set, requiring no gradient updates for novel E3 ligase tasks.

\subsubsection{Episodic meta-learning}

Training follows the episodic meta-learning protocol of Snell et al.\ \cite{Snell2017}. At each iteration, an episode is constructed by sampling a 2-way activity classification problem (active vs.\ inactive) from the relevant source-E3 training pool, with $K$ support samples and $Q$ query samples per class. \rev{The class labels within an episode are active/inactive, while E3 identity defines the meta-training/meta-test task split: source-E3 episodes are used to meta-train the encoder, and held-out target-E3 support/query episodes define the meta-test tasks.} The encoder processes all $2(K+Q)$ molecules, prototypes are estimated from the $2K$ support embeddings (Eq.~\ref{eq:prototype}), and the cross-entropy loss over the $2Q$ query predictions is minimized:
\begin{equation}
\mathcal{L} = -\sum_{(\mathbf{x}_q,\,y_q)\in Q}
  \log P(y_q\mid\mathbf{x}_q).
\label{eq:loss}
\end{equation}
By sampling diverse episodes throughout training, the encoder is optimized to produce embeddings in which within-episode class prototypes are well separated---\rev{a representational property intended to support prototype-based few-shot prediction when the E3-ligase-defined task changes at test time.}

\rev{For evaluation, episodes are drawn exclusively from held-out target E3 ligases (VHL, CRBN, or a rare ligase, depending on the transfer setting) that were never observed during the corresponding source-ligase training run. Target-ligase support compounds are used only to compute active/inactive prototypes, and target-ligase query compounds are used only for evaluation. In the bidirectional CRBN/VHL fixed-query benchmarks, target episodes use $K=2$ support samples per class with either $Q=3$ or $Q=5$ query samples per class. In the rare-E3 benchmark, each held-out ligase/seed episode uses $K=1$ support sample per class (one active and one inactive compound), followed by an all-remaining query protocol in which all eligible compounds from the same ligase that are not selected as support are used as query candidates. Thus, the rare-E3 benchmark contains one episode per rare ligase per seed, yielding five ligases $\times$ three seeds = 15 episodes before common-query coverage matching across baselines. Additional K/Q sensitivity analyses are provided in the Supplementary Information. This strict cross-ligase partition ensures that reported performance metrics reflect transfer to target-ligase chemical space rather than interpolation within the source-ligase chemical space.}

\subsection{Evaluation metrics}

Performance was evaluated using five complementary metrics: accuracy (ACC), macro F1-score (F1), balanced accuracy (BALACC), AUROC, and AUPRC. AUROC and AUPRC are designated as the primary metrics throughout, as they are threshold-independent and robust to class imbalance \cite{He2009}. Accuracy, BALACC, and F1 are reported as secondary metrics for reference; under the class-imbalanced episodic setting, macro F1 may favour methods with a tendency toward majority-class prediction and should be interpreted with caution. For episodic experiments, each metric was averaged across all tasks and episodes, and over three independent random seeds; results are reported as the mean and standard deviation.

\subsection{Implementation details}

All models were implemented in Python 3.9 using PyTorch \cite{pytorch2019} 2.0 and PyTorch Geometric 2.3. Molecular graph construction and 512-bit ECFP4 fingerprint generation (Morgan radius~$=2$) were performed using RDKit 2023.03 \cite{RDKit}. For ProMeta inference, the learned and fingerprint feature blocks were independently L2-normalized and concatenated using a fingerprint weight of 0.5. The GNN encoder comprised $L=3$ message-passing layers with hidden dimension $d=128$, followed by global mean pooling. Parameters were optimized using Adam \cite{adam2014} ($\beta_1=0.9$, $\beta_2=0.999$) with a learning rate of $1\times10^{-3}$ and weight decay of $1\times10^{-5}$. Hyperparameters were selected using the validation partition containing 20\% of the corresponding source-ligase POI targets, withheld from the training pool.

\rev{For each transfer direction, episodic meta-training tasks were sampled exclusively from the corresponding source-E3 molecules: CRBN for CRBN$\to$VHL and VHL for VHL$\to$CRBN ($K=2$, $Q=3$ per class during meta-training). The meta-trained encoder was then evaluated on support/query episodes from the held-out target E3 at meta-test. Evaluation configurations varied by benchmark and are described in Section~\ref{sec:experiments}. In each direction, the target E3 ligase was completely excluded from meta-training.} All evaluations used the same predefined support/query split-generation protocol and fixed seeds (42, 2025, and 3407); within each direction and K/Q setting, all compared methods received identical support/query rows to ensure a fair and reproducible comparison.

\subsection{Baseline evaluation protocol}
\rev{A key methodological challenge in evaluating cross-ligase few-shot generalization is that existing PROTAC activity predictors were not originally designed for cross-ligase support/query inference. To enable comparison within this setting, we re-evaluated or adapted all baselines in our pipeline using shared support/query splits.} Our baselines include PROTAC-STAN~\cite{chen2025}, DegradeMaster~\cite{liu2025}, a support-trained supervised GNN~\cite{Kipf2016}, and an ECFP feature representation~\cite{Rogers2010} coupled to a random forest~\cite{Breiman2001}. \rev{We emphasize that these baselines were developed for different prediction settings: in particular, DegradeMaster targets general supervised or semisupervised PROTAC prediction with 3D E(3)-equivariant modeling, whereas ProMeta is designed for cross-ligase few-shot prediction.} \rev{For RF + ECFP, a 500-tree RF is trained only on source-ligase ECFP4, RDKit5, and POI--E3 ACC features. Each compound is then represented by the frozen vector of per-tree positive-class probabilities concatenated with ECFP4, and an L2-regularized logistic head is fitted using only the target support set.} For the supervised GNN baseline, model parameters are randomly initialized and trained exclusively on the support set for each episode, with POI and E3 ligase sequence features incorporated via element-wise fusion. For PROTAC-STAN and DegradeMaster---which do not natively support this E3-separated episodic protocol---we retrain each model on the corresponding source-ligase training pool following its original procedure and default hyperparameters, then adapt the classifier head to each episode's support set via 50 gradient steps using AdamW, with all backbone parameters frozen. Query labels are used only for metric calculation. For each episode, the same available support and query rows---defined by shared split files generated under three random seeds (42, 2025, and 3407)---were supplied to every method; common-query subsets were used when an external model lacked complete input coverage. ProMeta instead concatenates the learned representation with the fixed ECFP4 block and computes class prototypes directly from target support compounds, requiring no encoder update at inference. Executable implementation details and frozen split files are supplied in the reproducibility repository.

\rev{As part of the ablation analyses, we further isolated the effect of the decision rule from the effect of the learned representation. In this matched-head analysis, GNN and PROTAC-STAN embeddings and DegradeMaster E(3)-equivariant latent features were evaluated with the same support-set prototype classifier; ECFP4 fingerprints were evaluated with nearest-centroid inference, and ProMeta was also evaluated with a support-trained linear head. DegradeMaster latent-feature results use the subset for which its required inputs are available. This reciprocal comparison tests whether ProMeta's performance is attributable only to non-parametric prototype inference or to the combination of episodic representation learning and support-based prototype construction.}

\section{Results}\label{sec:experiments}

\begin{figure*}[t]
\centering
\includegraphics[width=\textwidth]{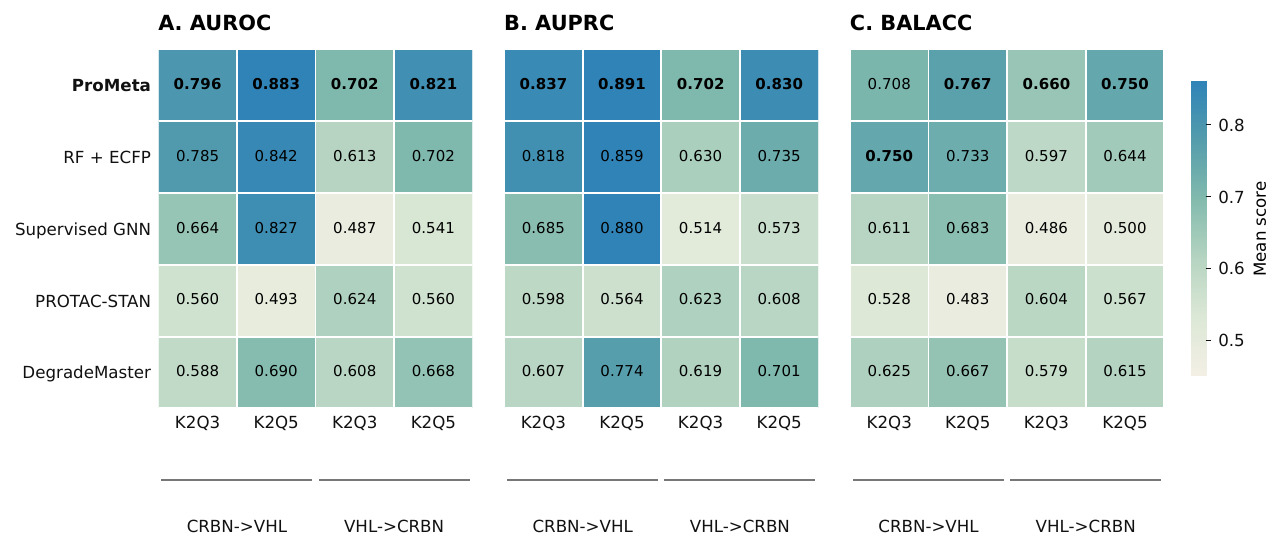}
\caption{\rev{Bidirectional CRBN/VHL transfer performance under strict episodic evaluation. Heatmaps show mean (\textbf{A})~AUROC, (\textbf{B})~AUPRC, and (\textbf{C})~balanced accuracy (BALACC) across three random seeds for the CRBN-to-VHL and VHL-to-CRBN $K=2, Q=3$ and $K=2, Q=5$ settings. The same predefined support/query split-generation protocol was used throughout, and all methods received identical support/query rows within each direction and K/Q setting. Darker colors indicate higher scores, and bold cell values indicate the best-performing method within each transfer setting and metric. Complete mean~$\pm$~standard deviation values are provided in Supplementary Table~S2.}}
\label{fig:bidirectional_crbn_vhl}
\end{figure*}

\subsection{\rev{Bidirectional cross-ligase transfer between CRBN and VHL}}

We first evaluate all methods on the CRBN to VHL transfer benchmark, in which models trained exclusively on CRBN-associated PROTAC molecules are assessed on episodic tasks constructed from VHL-associated molecules. Scaffold analysis confirmed zero overlap between the CRBN training set and both evaluation sets, with cross-set Tanimoto similarity substantially lower than within-CRBN similarity ($0.145$ and $0.163$ vs.\ $0.289$; Supplementary Fig.~S2), confirming that both benchmarks present genuine out-of-distribution generalization challenges at the molecular level. UMAP visualization of the learned molecular embeddings further reveals that CRBN and VHL compounds occupy largely distinct regions of the embedding space (Supplementary Fig.~S3), with degradation-active compounds forming loosely separable clusters within each ligase context, as further confirmed by VHL-specific embedding analysis (Supplementary Fig.~S4). This setting directly probes out-of-distribution generalization across E3 ligases, a central challenge in computational PROTAC modeling.

\rev{In this primary CRBN-to-VHL benchmark (Fig.~\ref{fig:bidirectional_crbn_vhl}), ProMeta retained the main conclusion of the original experiment. Under $K=2, Q=3$, ProMeta achieved an AUROC of $0.796$, compared with $0.785$ for RF + ECFP and $0.664$ for the supervised GNN baseline. Under $K=2, Q=5$, ProMeta achieved the highest AUROC of $0.883$, while RF + ECFP achieved $0.842$. These results indicate that a representation learned from CRBN episodes can support few-shot prediction on VHL using only a small VHL support set.}

\rev{To test whether this result was specific to one transfer direction, we additionally evaluated the reverse VHL-to-CRBN setting under the same strict cross-ligase episodic protocol. Reverse-transfer performance remained lower than CRBN-to-VHL transfer, consistent with the smaller VHL source set and the larger, more diverse CRBN target space. ProMeta achieved an AUROC of $0.702$ under $K=2, Q=3$ and $0.821$ under $K=2, Q=5$, compared with $0.613$ and $0.702$, respectively, for RF + ECFP. Thus, the bidirectional benchmark preserves the original CRBN-to-VHL interpretation while showing that few-shot cross-ligase generalization is empirically direction- and data-regime-dependent under substantial distribution shift. Complete mean~$\pm$~standard deviation values are provided in Supplementary Table~S2.}

\rev{To assess whether the stronger CRBN-to-VHL direction was attributable only to the larger CRBN source pool, we performed an independent 10-seed full-source versus VHL-size-matched retraining control. Under $K=2, Q=3$, the full-source and size-matched conditions achieved AUROCs of $0.731 \pm 0.079$ and $0.715 \pm 0.104$, respectively; under $K=2, Q=5$, the corresponding values were $0.883 \pm 0.044$ and $0.869 \pm 0.081$. These descriptive sensitivity results are reported in Supplementary Table~S3 and indicate that matching the source-pool size did not produce a consistent performance collapse.}

\rev{Additional K/Q sensitivity analyses, including larger fixed-query and all-remaining-query variants, are reported in Supplementary Table~S4.}

\begin{figure*}[t]
\centering
\includegraphics[width=\textwidth]{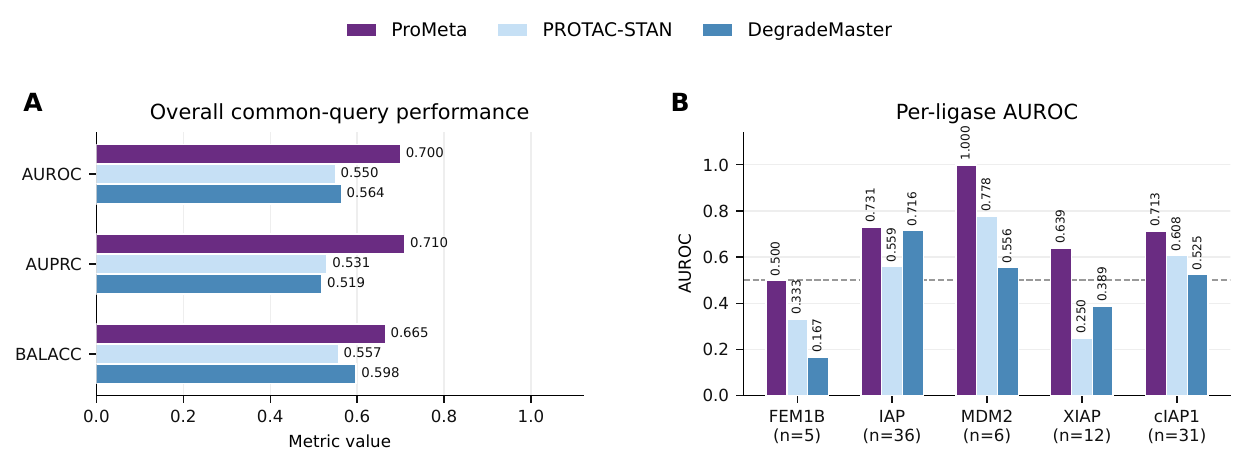}
\caption{
\rev{Rare-E3 common-query evaluation under the $K=1$ all-remaining-query protocol. (\textbf{A}) Overall AUROC, AUPRC, and balanced accuracy (BALACC) across 15 held-out rare-E3 episodes, constructed with one active and one inactive support compound per episode, and 90 strict common-query label-agree rows after coverage matching. (\textbf{B}) Per-ligase AUROC, with $n$ indicating the number of strict common-query label-agree rows after coverage matching. Dashed lines indicate the random AUROC reference of 0.5. For individual ligase--method pairs with very small query pools, an AUROC below 0.5 can result from only a few reversed positive--negative rankings and is not interpreted as evidence of stable inverse prediction. Full per-ligase AUROC/AUPRC/BALACC values are provided in Supplementary Table~S5.}}
\label{fig:rare_e3_bar}
\end{figure*}

\subsection{Cross-ligase transfer: CRBN to rare E3 ligases}

\rev{After evaluating bidirectional CRBN/VHL transfer, we next examine a more challenging rare-E3 setting with extremely limited labeled data. We construct a benchmark comprising five low-frequency ligases (FEM1B, IAP, MDM2, XIAP, and cIAP1), each with only a small number of labeled compounds in the curated dataset. This setting reflects a realistic deployment scenario in which only a few experimentally labeled compounds are available for novel ligase targets. For each rare ligase and random seed, the support set contains $K=1$ compound per activity class (one active and one inactive compound), and all remaining eligible compounds from the same ligase are used as query candidates after support selection rather than sampling a fixed numeric $Q$. This yields one episode per ligase per seed, or 15 held-out rare-E3 episodes across five ligases and three seeds. To ensure matched comparison across baselines, we report a strict common-query evaluation covering these 15 episodes and 90 strict common-query label-agree rows after coverage matching. The rare-E3 counts in Table~\ref{tab:e3_distribution} denote candidate labeled pools before episodic support/query construction, whereas the query rows reported here are episode-level evaluated rows after support selection and common-query coverage matching; unique query counts and reuse rates are audited in Supplementary Table~S6.}

As this benchmark targets cross-ligase generalization under extreme data scarcity, we focus on methods that can be applied to new E3-defined tasks with minimal supervision. Methods like RF and GNN trained from scratch are not well-suited to this setting, as the limited support samples provide insufficient signal to reliably learn task-specific decision boundaries. Therefore, we compare against PROTAC-STAN and DegradeMaster, two representative PROTAC degradation activity predictors evaluated here under the rare-E3 episodic protocol.

\rev{As shown in Fig.~\ref{fig:rare_e3_bar}, ProMeta achieves the best aggregate rare-E3 common-query performance, with an AUROC of $0.700$ across the 90 strict common-query label-agree rows. Under the same coverage-matched evaluation, PROTAC-STAN obtains an AUROC of $0.550$, and DegradeMaster obtains an AUROC of $0.564$. These results indicate that support-set prototype construction can preserve useful discriminative structure even when target-ligase labels are scarce, although the margin over adapted baselines remains modest in this extremely low-data setting.}

\rev{The complete per-ligase AUROC values and query-row counts are provided in Supplementary Table~S5, and the corresponding episode-level confidence intervals and support/query reuse audit are provided in Supplementary Table~S6. The per-ligase results are heterogeneous, but ProMeta achieves the highest AUROC across all five rare ligases. Some adapted-baseline point estimates fall below the random AUROC reference of 0.5. This behavior should not be interpreted as a stable tendency to predict the opposite biological label: AUROC is a pairwise ranking statistic, and in the smallest groups---for example, FEM1B with 5 episode-level query rows representing only 3 unique query compounds, and XIAP with 12 rows representing 6 unique compounds---one or two reversed positive--negative rankings can change the estimate substantially. Moreover, under $K=1$, the support-conditioned decision rule is highly sensitive to which single active and inactive compounds define the episode, and query compounds may recur across the three seeds. The below-0.5 values therefore reflect unstable realized rankings under an extreme one-shot stress test and should be read together with the confidence intervals and reuse audit in Supplementary Table~S6. Because several rare-E3 groups contain only a few common-query rows, we interpret this benchmark as an informative low-data stress test rather than as evidence for broad rare-ligase superiority.}


\subsection{Ablation study}

\rev{To isolate the contributions of episodic training and the support-set prototype decision rule, we expanded the original ablation beyond the previous GNN/GNN+MAML/ProMeta comparison. The revised ablation includes ProMeta, ProMeta + linear head, ProMeta without episodic training, supervised GNN encoder + prototype head, and ECFP4 nearest-centroid controls. Table~\ref{tab:ablation_main} summarizes the core mean AUROC results averaged over the two $K/Q$ settings within each transfer direction. Consistent with Supplementary Tables~S2 and S7, all $K=2, Q=3$ and $K=2, Q=5$ evaluations use the same predefined support/query split-generation protocol and the same three split seeds; within each direction and K/Q setting, all compared methods receive identical support/query rows. Full per-setting mean~$\pm$~standard deviation values and additional matched-head controls are provided in Supplementary Tables~S7 and S8.}

\begin{table}[tbp]
\centering
\caption{\rev{Core ablation results. Each cell reports mean AUROC averaged over the $K=2,Q=3$ and $K=2,Q=5$ settings.}}
\label{tab:ablation_main}
\scriptsize
\setlength{\tabcolsep}{3pt}
\begin{tabular}{lcc}
\toprule
Method & CRBN$\to$VHL & VHL$\to$CRBN \\
\midrule
ProMeta & \textbf{0.840} & \textbf{0.761} \\
ProMeta + linear head & 0.814 & 0.691 \\
ProMeta w/o episodic training & 0.818 & 0.554 \\
GNN encoder + prototype head & 0.806 & 0.520 \\
ECFP4 nearest-centroid & 0.799 & 0.689 \\
\botrule
\end{tabular}
\end{table}

\rev{The comparison with ProMeta without episodic training shows a direction-dependent contribution. Removing episodic training changed mean AUROC from $0.840$ to $0.818$ in CRBN-to-VHL transfer and from $0.761$ to $0.554$ in VHL-to-CRBN transfer. The larger reverse-transfer gap indicates that the benefit of episodic representation learning is not uniform across data regimes.}

\rev{The matched-head controls further show that the prototype decision rule is beneficial but not sufficient by itself. The ProMeta + linear head experiment provides the direct classification-head control. Replacing prototype inference with a support-trained linear head reduced mean AUROC in both transfer directions, while applying prototype inference to supervised GNN embeddings or ECFP4 fingerprints also remained below ProMeta. Taken together, these results indicate that neither episodic training nor prototype inference alone fully accounts for the observed cross-ligase performance. Instead, the combination of episodic representation learning with a support-set, non-parametric prototype decision mechanism provides a suitable inductive bias for the evaluated data-scarce transfer setting.}

\begin{figure*}[tp]
\centering
\includegraphics[width=\textwidth]{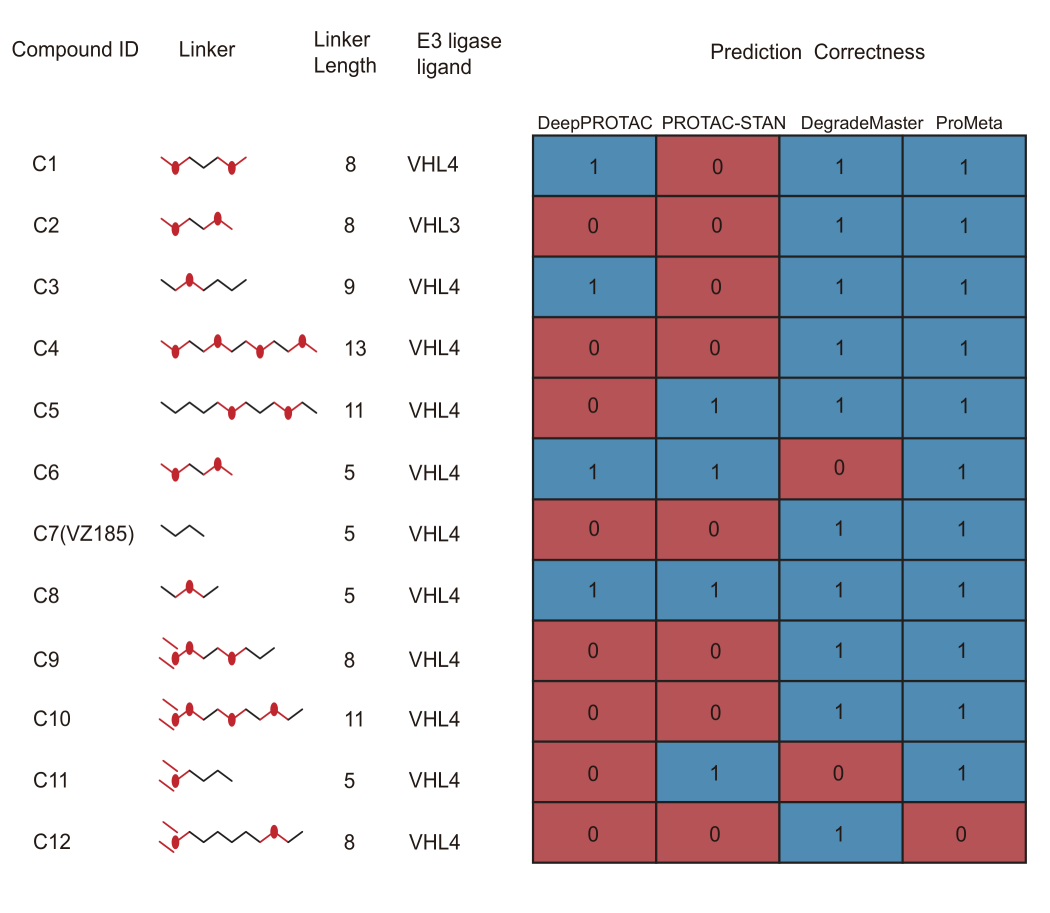}
\vspace{-2em}  
\caption{
Degradability prediction of VZ185-derived PROTAC candidates \cite{Zoppi2019}. Compounds C13--C16 lack a POI-binding warhead and were excluded; C1--C12 were used for evaluation. Each cell indicates whether a model's prediction matches the experimental label (1~=~correct, 0~=~incorrect). Heatmap columns correspond to DeepPROTAC, PROTAC-STAN, DegradeMaster, and ProMeta.}
\label{fig:case_heatmap}
\end{figure*}

\subsection{Case study: VZ185-derived PROTAC candidates}

Following the evaluation protocol of DeepPROTAC \cite{Li2022}, we assessed the practical utility of ProMeta beyond episodic benchmarks in a case study of 12 structurally related PROTAC candidates derived from the VZ185 scaffold \cite{Zoppi2019}. These compounds all recruit VHL and share the same warhead and E3 ligase ligand substructures, differing mainly in linker composition. This controlled setting examines each model's sensitivity to linker-associated structure--activity relationships. As shown in Fig.~\ref{fig:case_heatmap}, ProMeta achieves 91.7\% classification accuracy (11 out of 12 compounds), outperforming DeepPROTAC, PROTAC-STAN, and DegradeMaster on this small retrospective task. \rev{To examine performance beyond the VZ185 series, we additionally evaluated two independent retrospective medicinal-chemistry series using the prespecified main-experiment checkpoint family. ProMeta correctly classified 11 of 14 CRBN-recruiting CDK6 degraders (78.6\%) and all 6 VHL-recruiting BCL-xL degraders (100.0\%) \cite{Su2019, Zhang2020}; compound-level experimental annotations and predictions are provided in Supplementary Table~S9 and the reproducibility files. Given their small sample sizes, these additional series provide complementary scaffold-level evidence and should not be interpreted as evidence of broad chemical-series superiority.} ProMeta correctly classifies several active VZ185-series compounds that all three baseline models misclassify, which is consistent with sensitivity to linker-associated structural variation. Linker length and geometry are known to influence ternary-complex formation and the spatial positioning required for productive ubiquitin transfer \cite{Gadd2017, Zoppi2019}; the present retrospective result suggests, but does not establish mechanistically, that ProMeta may capture related structure--activity signals.

\section{Conclusion and discussion}

\rev{ProMeta addresses a specific limitation that is not the primary focus of most existing PROTAC activity predictors: cross-ligase degradation prediction under limited labeled data. Unlike conventional supervised approaches, ProMeta reformulates this setting as a few-shot learning problem for cross-ligase generalization and predicts target-ligase query compounds by constructing active/inactive prototypes from a small target support set without updating model parameters. The revised bidirectional CRBN/VHL experiments show that ProMeta is not limited to the original CRBN-to-VHL direction, but also reveal that cross-ligase few-shot generalization is direction- and data-regime-dependent under substantial distribution shift. The K/Q sensitivity and classification-head controls further indicate that the conclusion is not driven by a single support/query split or by a conventional linear head.}

A notable observation is that RF + ECFP, despite relying largely on 2D fingerprints, achieves competitive AUROC values in selected settings. This result indicates that representation complexity alone does not determine performance under the evaluated protocol and that source--target distribution mismatch remains important. CRBN and VHL recruit structurally distinct chemical scaffolds and engage different POI distributions, which provides a plausible chemical and biological basis for the observed direction dependence; the present predictive experiments do not, however, isolate a single mechanistic cause. \rev{The lower performance of PROTAC-STAN and DegradeMaster in our episodic evaluation should therefore be interpreted in light of their original objectives: they were designed for general supervised or semisupervised PROTAC prediction, with DegradeMaster further emphasizing 3D E(3)-equivariant modeling, rather than for cross-ligase few-shot support/query prediction. Thus, our results do not claim general superiority over these models in their native prediction setting; instead, they show that architectures optimized for data-rich prediction do not automatically solve cross-ligase few-shot generalization.}

Despite its advantages, ProMeta currently relies on complementary 2D molecular graph and ECFP4 fingerprint representations together with protein-sequence context, without explicit 3D conformational or ternary-complex information. \rev{Lightweight 3D augmentation controls and an intersection-trained, coverage-limited diagnostic comparison with the geometry-aware SE(3)-PROTACs baseline \cite{Kothakapu2026} are provided in Supplementary Tables~S10 and S11, respectively.} In addition, the available source-ligase data remain concentrated in CRBN and VHL, and the rare-E3 analyses are necessarily limited by small query sets. Therefore, the cross-ligase generalization evidence should be interpreted within the controlled few-shot support/query evaluation used here rather than as exhaustive validation across all E3 ligase families. Future work will incorporate E(3)-equivariant encoders and structure-based ternary complex features, and extend meta-training to additional source-ligase contexts as PROTAC-DB continues to grow.

\rev{In summary, we present ProMeta, a prototype-based framework for cross-ligase few-shot PROTAC degradation prediction. Across the evaluated protocols, ProMeta provides a practical strategy for predicting held-out E3 ligase contexts with minimal labeled support data, with the strongest evidence in CRBN-to-VHL transfer and more heterogeneous but informative behavior in the reverse VHL-to-CRBN and rare-E3 settings. These results support ProMeta as a few-shot framework for cross-ligase generalization.} As therapeutically relevant E3 ligases continue to expand beyond well-characterized cases, methods that can generalize with minimal labeled data will be important for accelerating next-generation PROTAC discovery.

\section{Supplementary Data}

Supplementary data are available with this preprint.


\section{Conflict of interests}
No conflict of interest is declared.


\section*{Acknowledgements}
This study was supported by Yuelushan Laboratory Breeding Program (No. YLS-2025-ZY03024); the National Natural Science Foundation of China (Grant No. 62372159, 32400506); the Natural Science Foundation of Hunan Province (Grant No. 2024JJ4008); Fundamental Research Funds for the Central Universities (Grant No. 541109030062).



\bibliographystyle{unsrt}
\bibliography{reference}


\end{document}